\documentclass[10pt,twocolumn,letterpaper]{article}

\usepackage{iccv}
\usepackage{times}
\usepackage{epsfig}
\usepackage{graphicx}
\usepackage{amsmath}
\usepackage{amssymb}
\usepackage{bm}
\usepackage{multirow} 
\usepackage{booktabs}
\usepackage[export]{adjustbox}
\usepackage{mathtools, nccmath}
\usepackage{caption}
\usepackage{subcaption}
\usepackage{fancyhdr}
\DeclarePairedDelimiter{\nint}\lfloor\rceil

\usepackage[breaklinks=true,bookmarks=false]{hyperref}

\iccvfinalcopy 

\def\iccvPaperID{49} 

\ificcvfinal\fi

\begin{document}
\setlength{\headheight}{15.2pt}


\title{Efficient Neural PDE-Solvers using Quantization Aware Training}

\author{Winfried van den Dool$^1$, Tijmen Blankevoort$^2$, Max Welling$^1$, Yuki M. Asano$^1$\\
$^1$QUVA Lab at the University of Amsterdam, $^2$Qualcomm AI Research\footnotemark \\
{\tt\small w.v.s.o.vandendool@uva.nl}
}

\maketitle
\ificcvfinal\thispagestyle{empty}\fi

\begin{abstract}
   In the past years, the application of neural networks as an alternative to classical numerical methods to solve Partial Differential Equations has emerged as a potential paradigm shift in this century-old mathematical field. However, in terms of practical applicability, computational cost remains a substantial bottleneck. Classical approaches try to mitigate this challenge by limiting the spatial resolution on which the PDEs are defined. For neural PDE solvers, we can do better: Here, we investigate the potential of state-of-the-art quantization methods on reducing computational costs. 
   We show that quantizing the network weights and activations can successfully lower the computational cost of inference while maintaining performance. Our results on four standard PDE datasets and three network architectures show that quantization-aware training works across settings and three orders of FLOPs magnitudes.
   Finally, we empirically demonstrate that Pareto-optimality of computational cost vs performance is almost always achieved only by incorporating quantization. 
\end{abstract}
\renewcommand*{\thefootnote}{\fnsymbol{footnote}}
\footnotetext{$^*$an initiative of Qualcomm Technologies, Inc.}
\renewcommand*{\thefootnote}{\arabic{footnote}}
\setcounter{footnote}{0}

\section{Introduction}
In  many scientific fields, mathematical models of observed phenomena are expressed in terms of Partial Differential Equations, with the solution commonly represented by a function of a single time variable and one or more spatial variables.
While many PDEs can be described compactly (\eg the famous Schrödinger's equation from quantum physics, or Navier-Stokes' equations from fluid dynamics) it is usually impossible to write down explicitly the formulas for their exact solutions. Rather, there is a vast amount of scientific research on methods to numerically approximate solutions.

When it comes to practical applications, computational cost and available resources play a significant role in this field of research. For most methods a natural trade-off occurs where one may use available resources to either speed up the computation, or increase the resolution on which the PDE is defined, leading to a slower but more accurate result. This latter strategy is of particular interest for PDEs that are highly non-linear and potentially exhibit chaotic behavior. Small eddies in a turbulent fluid may, for example, be overlooked when the spatial resolution is too low, leading to increasing errors in the solution function over time.
One of the most important fields where computational resources form a bottleneck is climate prediction. The challenge of determining how many degrees the planet will warm over the next decades is a computational one, with supercomputers running massive ensemble-type methods. Insufficient resolution is quite often the main cause for sub-optimal modeling accuracy~\cite{randall2007climate, scale2, kurth2023fourcastnet}. Indeed, over time climate models have become better for a large part simply because the resolution of the grid that they are defined on was allowed to increase, due to more efficient computation and more available resources~\cite{schneider2017climate,climate1, climate2}.

It is precisely in this context of the computational cost of the traditional solvers that neural PDE solvers are becoming an interesting alternative. A single forward pass in a neural network is extremely fast compared to the iterative solving procedure of classical methods, and most of the operations are large matrix multiplications that can be easily handled by GPUs. While a neural network does need training first, possibly requiring data generated by classical solvers, its efficiency benefit comes from ``recycling'': After training is complete, the neural network has learned to generalize to different initial conditions, whereas conventional solvers only solve the PDE for one specific configuration of initial conditions at a time. In practice, a deep learning weather model would have to be trained only once, but could then be applied every day to predict the next day's weather. 
Furthermore, when training data is sufficiently abundant in the real world, using that data directly comes with the added advantage that complex phenomena do not need to be first accurately modeled. Indeed, neural PDE solvers trained on real weather data alone have shown promising results~\cite{Pangu,realweather2,kurth2023fourcastnet}.

Our goal is not to improve on state-of-the-art models in the conventional sense. Rather, our starting point is to borrow from existing networks and architectures and focus on their reducing their inference cost. With neural PDE solvers becoming mainstream for practical use, this measure becomes more important than test loss alone. Our contribution is in finding the optimal way to deal with the loss and computational cost trade-off. 
There are several orthogonal approaches to achieve lower computational costs. On the one hand, as in classical PDE solvers, the foremost way to make computation manageable is to reduce the resolution on which the PDE is defined. On the other hand, deep learning researchers have developed a variety of different techniques specifically for neural networks, most notably quantization, the focus of this work.
Overall, we make the following contributions:
\begin{itemize}
\setlength\itemsep{0.2em} 
\item We provide an evaluation of 3 neural network-based PDE solvers \cite{PDEArena1, FNO, GalerkinTransformer} under 5 weight quantization scenarios. We use state-of-the-art quantization methods to provide exemplary benchmarks on 4 of the most common datasets.
\item We are the first to investigate both spatial resolution and model (weights and activations) resolution simultaneously as hyper-parameters in designing neural PDE-solvers. We develop a hybrid approach of applying quantization and modifying spatial resolution depending on the problem to achieve compute- and accuracy-optimal results.
\item We provide an extensive analysis of the trade-off between computational cost and errors for different quantized models and demonstrate that a certain level of quantization is almost always necessary to be Pareto-optimal on the accuracy-cost curve.
\end{itemize}

\section{Related Works}

\noindent \textbf{Neural PDE solvers.} There are already many deep learning methods  that aim to solve PDEs \cite{GalerkinTransformer,UNet1,PDEArena1,PDEArena2, mpgnn, raissi2019physics, huang2022partial}. We base our research on the most common neural PDE surrogate architectures, namely the Fourier Neural Operator \cite{FNO}, UNet \cite{UNet1,PDEArena1} and a type of Transformer \cite{GalerkinTransformer}.
In terms of datasets, there are two recent larger-scale attempts at proper benchmarking of the field, namely PDEArena \cite{PDEArena1} and PDEBench \cite{PDEBench}; from both of which we borrow our training and evaluation datasets. 

\vspace{1em}
\noindent  \textbf{Quantization.} With regards to quantization methods we rely on AIMET \cite{AIMET}\footnote{AIMET is a product of Qualcomm Innovation Center, Inc. (BSD-3)}, which provides state-of-the-art quantization techniques. In particular, we utilize Quantization Aware Training \cite{jacob2018quantization, jain2020trained}, utilizing the straight-through estimator \cite{straightthroughestimator} to approximate gradients of rounding operators. We allow quantization ranges to be trainable parameters as well, as in \cite{esser2020learned}.

\vspace{1em}
\noindent  \textbf{Low precision PDE solving.} Recently, there has been increasing interest in running classical methods for solving climate models with lower precision as well. In \cite{Kimpson_2023} reduced precision of Float32 and Float16 has been used, as opposed to the standard precision of Float64. However, these solvers are based on traditional approaches, and not on neural networks.

In contrast to previous PDE surrogates, i.e. (deep learning) functions that aim to approximate the original PDE solution, we especially care for computational cost. While it is generally investigated that quantization decreases the accuracy of deep learning models as well as that standard PDE solvers' accuracy depends on their grid size, we explicitly compare both techniques in terms of how much they reduce computational costs. To make this possible we take a particular interest in the cost-loss trade-off of both techniques separately as well as their combined effects. Also, while a classification algorithm may exhibit a clear breaking point at which it can no longer accurately predict the majority of correct labels, the prediction of a PDE solution is measured in terms of (mean squared) errors between functions. As such there is no clear definition of what a ``correct" prediction is, and consequentially there is no clear measure of when a PDE surrogate is successful. This means that a wide regime of models of varying qualities becomes interesting to investigate, both in the low-cost as well as the low-loss regions of the cost-loss trade-off.


\section{Method} 

\subsection{Theoretic background and notations}
As we work with synthetic data, we are given numeric solutions to a given known PDE. Let $X\subset \mathbb{R}^d$ be a spatial domain and $[0,T]$ be a time window on which $\bm{u}:X\times [0,T] \rightarrow \mathbb{R}^n$ is the solution to the partial differential equation 
\begin{eqnarray}
\frac{\partial \bm{u}}{\partial t} = F\left(\bm{x},\bm{u},\frac{\partial \bm{u}}{\partial \bm{x}},\frac{\partial^2 \bm{u}}{\partial \bm{x}^2},\dots \right),
\end{eqnarray}
with initial condition $\bm{u}^0(\bm{x})$ at time $t=0$ and boundary conditions defined by the operator $\mathcal{B}[\bm{u}](t,\bm{x})=0$ when $\bm{u}$ is on the boundary $\partial X$ of the domain $X$. Here $F$ can be any function, though we specify the precise form for different datasets in the Supplementary Information (SI).
Assume that at some time $t$, $\bm{u}(t,\bm{x})$, as well as prior values of $\bm{u}$, are known. Let $\tau>0$ be some step-size with $t+\tau <T$. We can then try to predict one time step into the future by
\begin{eqnarray}
\label{equation_timeintegral}
    \bm{u}(t+\tau,\bm{x}) &=& \bm{u}(t,\bm{x}) + \int_{t}^{t+\tau} \frac{\partial \bm{u}(t,\bm{x})}{\partial t} \mathrm{d}t.
    \\
    \nonumber 
    && \forall t \in [0,T-\tau], \bm{x} \in X
\end{eqnarray}
Given that theoretically the value of $\bm{u}(t,\bm{x})$ determines, through its spatial derivates and the PDE, all future values, a first attempt at defining a useful operator may be to replace the right-hand side of the above equation by an operator $\mathcal{F}_\tau$ such that
\begin{eqnarray}
    \bm{u}(t+\tau,\bm{x}) = \mathcal{F}_\tau[\bm{u}(t,\bm{x})] \quad \forall t \in [0,T-\tau], \bm{x} \in X,
\end{eqnarray}
indeed ignoring values of $\bm{u}(t',\bm{x})$ for $t'<t$. We are then interested in approximating this operator $\mathcal{F}_\tau$ by a neural network $\mathcal{G}$.
However, in practice the spatial and time domains are discretized on regular grids $\mathcal{X} \subset X$ and $\mathcal{T}\subset [0,T]$ with $|\mathcal{X}|=N_x$ and $|\mathcal{T}|=N_t$. Crucially, when $\bm{u}(t,\bm{x})$ is defined on $\mathcal{X}$ rather than the full domain $X$, its spatial derivatives are no longer determined, breaking the connection from $\bm{u}(t,\bm{x})$ to $\frac{\partial \bm{u}}{\partial t}$ through the PDE $F$. Consequentially, the full future cannot be exactly determined from $\bm{u}(\bm{x},t)$ defined on $\mathcal{X}$ alone.
Note that even knowing the spatial derivatives exactly on the grid $\mathcal{X}$ at time $t$ is not enough, as they are required at all later times as well if we are to solve the integral in equation \ref{equation_timeintegral}.
Now the values of $\bm{u}(\bm{x},t')$ for $t'<t$ may still provide helpful information, and we may want to include those as inputs in our neural network approximation of $\mathcal{F}$. One can think for example that such values may allow the backward difference approximation of $\partial \bm{u}(t,\bm{x})/\partial t$ as an alternative to using the PDE $F$ with the approximated spatial derivatives.

Incorporating multiple past time steps, the neural network approximation is given by the operator $\mathcal{G}$, defined by
\begin{eqnarray}
    \bm{u}(t_n+\tau,\bm{x}) &=& \mathcal{G}_\tau[\left(\bm{u}(t_i,\bm{x})\right)_{i=1}^{n}] \quad \\
    \nonumber && \text{for} \; \; t_i,t_n+\tau \in \mathcal{T}, \bm{x} \in \mathcal{X}.
\end{eqnarray}
In what follows we assume $\bm{u}$ to be defined on $\mathcal{T} \times \mathcal{X}$.


\subsection{Training setup}
Let the true $\bm{u}(\bm{x},t)$ be defined everywhere on our grid $\mathcal{T}\times \mathcal{X}$. For each such trajectory $\bm{u}(\bm{x},t)$ in a mini-batch we first randomly select a starting time, after which we take a fixed set of subsequent input indices $\mathcal{T}_\text{input}\subset \mathcal{T}$, and target indices, $\mathcal{T}_\text{target}\subset \mathcal{T}$. We have $N_t-|\mathcal{T}_\text{input}|$ input steps available, but we are only training on $|\mathcal{T}_\text{target}|$ time steps. Therefore, to make full use of the available data we repeat each epoch $\lceil{(N_t-|\mathcal{T}_\text{input}|)/|\mathcal{T}_\text{target}|\rceil}$ times, selecting new random indices every time so that in expectation the full dataset is used every epoch.

For $|\mathcal{T}_\text{target}|>1$ outputs we may use either temporal bundling (letting $\mathcal{G}$ output multiple subsequent time steps in one forward pass, see for instance \cite{mpgnn}), or a recurrent approach, where the last outputs are fed to the network as new inputs, or a combination of both. 
Similar to \cite{mpgnn} we may apply \textit{pushforward}, meaning that we only backpropagate the loss through the last part of the target indices.

Let $\bm{u}(t,\bm{x})$ be the true targets and $\hat{\bm{u}}(t,\bm{x})$ the corresponding neural network predictions for $(t,\bm{x})\in\mathcal{T}_\text{target}\times\mathcal{X} $. The loss, assuming no pushforward is applied, is then defined by
\begin{eqnarray} \label{equation:loss}
    \frac{\sum_{x \in \mathcal{X}} \sum_{t\in \mathcal{T}_\text{target}}\sum_{i=1}^{N_\text{fields}} \|\bm{u}_i(\bm{x},t)-\hat{\bm{u}}_i(\bm{x},t)\|_2^2}{|\mathcal{X}| |\mathcal{T}_\text{target}| N_\text{fields}} ,
\end{eqnarray}
where $N_\text{fields}$ is the dimensionality of $\bm{u}(t,\bm{x})$.\\


\subsection{Quantization}
By default, the weights and activations of the neural network $\mathcal{G}$ are defined as floating point numbers using 32-bit precision. We can, however, store the weights and activations as integer values, to make (matrix) multiplications much more efficient. A floating-point weight matrix $\bm{W}$ can be expressed as a single scalar multiplied by a matrix of integer values, and a remainder term $\bm{\epsilon}$.
\begin{eqnarray*}
    \bm{W} = s_W \cdot \bm{W}_\text{int} + \bm{\epsilon}_W
\end{eqnarray*}
Similarly a vector of activations $\bm{v}$ may be expressed as
\begin{eqnarray*}
    \bm{v} = s_v \cdot (\bm{v}_\text{int}-\bm{z}_v) + \bm{\epsilon}_v,
\end{eqnarray*}
where we have added a zero-point $\bm{z}_v$ to allow assymetric quantization of the activations. Matrix-vector multiplication now becomes:
\begin{eqnarray*}
    \bm{W}\bm{v} &=& (s_W \cdot \bm{W}_\text{int} + \bm{\epsilon}_W)(s_v \cdot (\bm{v}_\text{int}-\bm{z}_v) + \bm{\epsilon}_v)\\
    &=& s_W s_v \bm{W}_\text{int} \bm{v}_\text{int} - s_W s_v \bm{W}_\text{int} \bm{z}_v + \text{error terms}
\end{eqnarray*}
As the second term depends only on the weight matrix $W$ and quantization parameters $s_W$, $s_v$ and $z_v$, it can be computed beforehand. If we then ignore the error terms, only the first term is remaining during inference: a matrix-vector multiplication of integer values. 
For a given bitwidth $b$, there are $2^b$ possible integer values to choose from. Given a zero-point $z$ and scale factor $s$, the quantization grid limits $q_\text{min}$ and $q_\text{max}$ are then determined by $-sz$ and $s(2^b-1-z)$. Any values lying outside this range will be clipped to its limits, incurring a \textit{clipping error}.
The full quantization function $q(\cdot)$ is given by
\begin{eqnarray}
    \bm{x}_\text{int}=q(\bm{x};s_x,\bm{z}_x,b)= \text{C}\left(\nint*{\frac{\bm{x}}{s_x}}+\bm{z}_x;0,2^b-1\right),
\end{eqnarray}
where $\nint{\cdot}$ is the round-to-nearest operator and $\text{C}$ is a clamping function defined as:
\begin{eqnarray}
    \text{C}(x;a,c) =\begin{cases}
			a, & x <a \\
            x, & a\leq x \leq c \\
            c, & x>c
		 \end{cases}
\end{eqnarray}

To get a quantized network we start with a pre-trained floating point network and then optimize for the values of $\bm{s}$ and $\bm{z}$ that minimize the error terms that result from both clipping and rounding errors. This happens independently per layer, using arbitrary dummy data or training data in the forward pass.
Next, we also apply Quantization Aware Training (QAT). This is a fine-tuning step, further training the quantized network using stochastic gradient descent on the original loss function. Training with quantized weights and activations is possible using the straight-through estimator \cite{straightthroughestimator}.\\
Noting that the majority of the computational cost comes from (matrix) multiplications, we only quantize the inputs and weights of neural network layers that are based on matrix multiplications, and let the outputs (together with the biases) be accumulated in floating point format.

\subsection{Changing resolutions}
We change resolutions using bilinear interpolation for the 2D datasets and linear interpolation for the 1D datasets. We leave experimentation with learnable resolution operators for future research. However, we have chosen (bi)linear interpolation deliberately for its speed, which is the prime reason to change resolutions in the first place.
If the network is defined on a different resolution than the data, we apply resolution-altering operators before and after it during both validation and training, thus also backpropagating losses through the resolution-altering operators.

\subsection{Model cost computation} \label{section: model cost computation}
The proxy we use for model inference efficiency is based on counting multiplication and addition operations per layer. For a given quantized network module with $M$ multiplication and $A$ addition operations, and integer bitwidths of $b_w$, $b_a$ for the weights and activations respectively, the cost is defined as
\begin{eqnarray*}
    M\cdot b_w \cdot b_a+A\cdot b_a,
\end{eqnarray*}
where multiplications are considered between weights and inputs, and addition operations only apply to outputs. (We do not quantize bias vectors.) 
The full network cost is the sum of the cost over all layers. 
If a network works on a different resolution, we also take the costs of altering the resolution before and after the forward pass into account.
Details on the number of multiplications and additions per layer are provided in the code.
We have observed no increased losses from quantizing any parameter to bitwidth 16 under any circumstance, and therefore assume that all floating point operations can be harmlessly replaced by their Int16 counterpart. This enables us to measure the computational cost of non-quantized operations as if they were Int16 operations and to generally use fixed-point integer operations as a measure throughout all our comparisons of different model inference costs.
To find the number of multiplication and addition operations in a given module we use \cite{deepspeed}, slightly adapted for our needs. The deepspeed model profiler lets you choose either FLOPs or MACs as a measure for module cost. We use the fact that one MAC operation consists of a single addition and multiplication and assume that the remaining operations are all addition-type in terms of complexity. Hence, when deepspeed outputs $X$ FLOPs and $Y$ MACs for a certain network module, we know that there are $Y$ multiplications and $X-Y$ remaining FLOPs that we consider additions. A few exceptions and special cases, as well as deliberate deviations from deepspeed, can be found in Appendix \ref{appendix:inferencecost}.

\section{Experiments}
We use three popular neural-surrogate architectures: FNO, UNet and Transformer, applied to datasets based on 4 PDEs: Diffusion-Sorption (1D),  Burgers' (1D), Navier-Stokes (2D) and Darcy (2D). The datasets are described in more detail in Appendix \ref{appendix:datasets}.
We show benchmark results of the original networks, as well as the results of the networks after quantization and scaling. For each result, we present the test loss as well as the computational cost of inference. We first compare quantization and reduction of spatial resolution as two orthogonal approaches to decrease computational cost, showing their impacts separately. Finally, we apply both methods simultaneously on higher-resolution versions of the datasets.

\subsection{Implementation}
For a given spatial resolution, we first train the floating point network in a regular fashion, i.e., aiming for the lowest possible loss. We use Adam \cite{adam} with weight decay and a cosine annealing learning rate scheduler \cite{cosannealing} with a linear warmup. Next, we quantize the network, first finding the best quantization parameters using AIMET's built-in optimization tool on $20\%$ of the training data. Then we fine-tune the quantized network, training again using Adam and cosine annealing with linear warmup, but without weight decay, with a smaller learning rate, and for fewer epochs. Specific parameters for training and fine-tuning are provided in Appendix \ref{appendix:params}, as well as the length of the input and output trajectories, i.e., the number of time steps used.
If a model that operates on a different resolution than the original data resolution is unrolled several times, i.e., when training and testing on a longer output trajectory, the resolution is only altered after the first input, and after the output just before the loss is computed: intermediate values that are fed back into the network are kept in the network resolution.
The quantization regimes used throughout the experiments are: [\texttt{w4a4}, \texttt{w4a8}, \texttt{w8a8}, \texttt{w8a16}], with \texttt{wXaY} refering to quantizing weights to bitwidth \texttt{X} and activations to bitwidth \texttt{Y}.\footnote{We found it better to use a higher activation bitwidth than weight bitwidth when using different bitwidths for weights and activations}. We do not quantize the first and last layer of each model, having observed that this improves performance with negligible impact on inference cost. 
Unless mentioned otherwise, we use scaling regimes, i.e., factors by which we scale input data resolution, of [$0.7, 0.5, 0.3, 0.2$]. For 2D data, we scale both dimensions by this factor, thus the actual number of spatial points depends quadratically on the scaling factor.

\subsection{Results}
To briefly introduce the models and datasets\footnote{Datasets were solely downloaded and evaluated by QUVA.} used we present an overview of the unquantized, standard-resolution models' loss and inference cost on all datasets in Tables \ref{table: results general loss} and \ref{table: results general cost}. Here we use the number of floating point operations (flops), as taken from deepspeed \cite{deepspeed}, directly as a measure of model inference cost. We note that, as there is no clear measure of when a PDE surrogate is successful (compared to \eg classification algorithms), a wide regime of varying MSE losses and costs is interesting to investigate. In Tables \ref{table: results general loss} and \ref{table: results general cost} we can see that the PDEs have vastly different possible solutions in terms of MSE and cost.

\begin{table}[]
\begin{tabular}{lrrr}
\toprule
\quad Val. MSE loss           & FNO              & UNet             & Transformer \\ 
\midrule
Diff-Sorp (1D)     & \textbf{4.12e-8} & 4.26e-8          & 4.33e-8     \\ 
Burgers' (1D)       & \textbf{5.62e-4} & 2.19e-3          & 1.40e-3      \\ 
Navier-Stokes (2D) & 2.85e-3          & \textbf{1.10e-3} & 5.36e-3     \\ 
Darcy (2D)         & 1.52e-2          & \textbf{6.92e-3} & 8.47e-3     \\ 
\bottomrule
\end{tabular}
\caption{\label{table: results general loss} \textbf{Overview of unquantized model performances.} We report Mean Squared Error (MSE) loss for the unquantized neural networks that we analyze in this paper. 
}
\end{table}

\begin{table}[]
\begin{tabular}{lrrr}
\toprule
\quad FLOPS (in millions)   & FNO    & UNet            & Transformer  \\
\midrule
Diff-Sorp (1D)     & 15 & 7          & {4} \\ 
Burgers' (1D)       & 21 & 14          & {9} \\ 
Darcy (2D)         & 150 & {34} & 69          \\ 
Navier-Stokes (2D) & 210 & 181          & {118} \\ 
\bottomrule
\end{tabular}
\caption{\label{table: results general cost} \textbf{Model FLOPS across different datasets.} In this paper, we evaluate how we can increase the efficiency of models that span almost three orders of magnitude in FLOP inference cost.
}
\end{table}
\subsubsection{Quantization compared to reducing resolution}
In what follows we refer to models being applied on reduced resolution as ``scaled" models, although it is actually their input data that is scaled. Implicitly such models are smaller as a result of them being applied on lower-resolution data, but we do not change hyperparameters like layer width or number of layers.
\begin{figure}[h]
    \centering
    \includegraphics[width=0.46\textwidth]{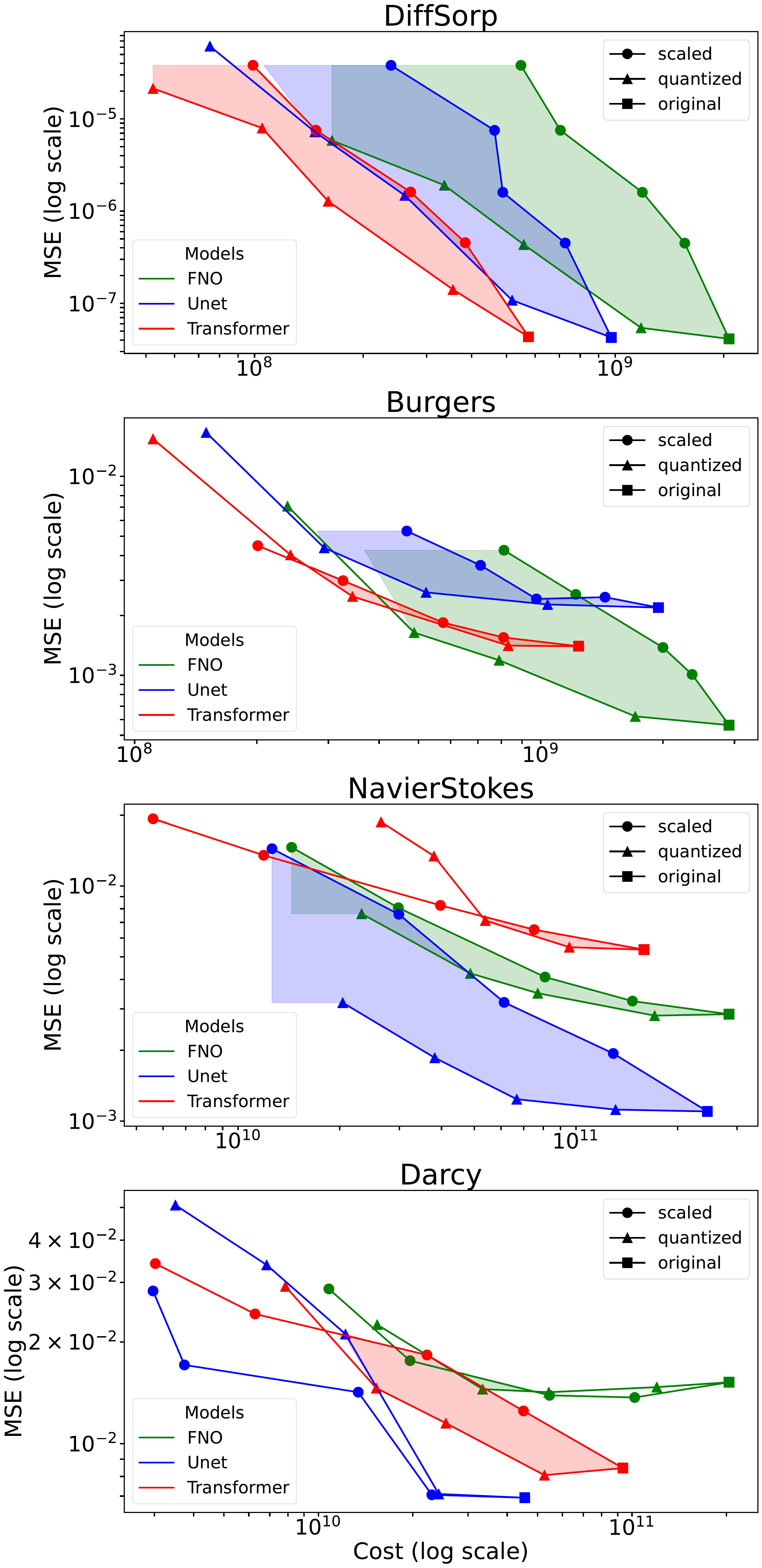}
    \caption{\textbf{Quantization for more robust cost-reduction}. We compare model quantization against scaling spatial resolution as ways to reduce compute costs. We show the cost in FLOPS versus the Mean Squared Error (MSE) across 4 PDE datasets and three model types. 
    }
    \label{fig:results1}
\end{figure}

\begin{figure}[t]
    \centering
    \includegraphics[width=0.4\textwidth]{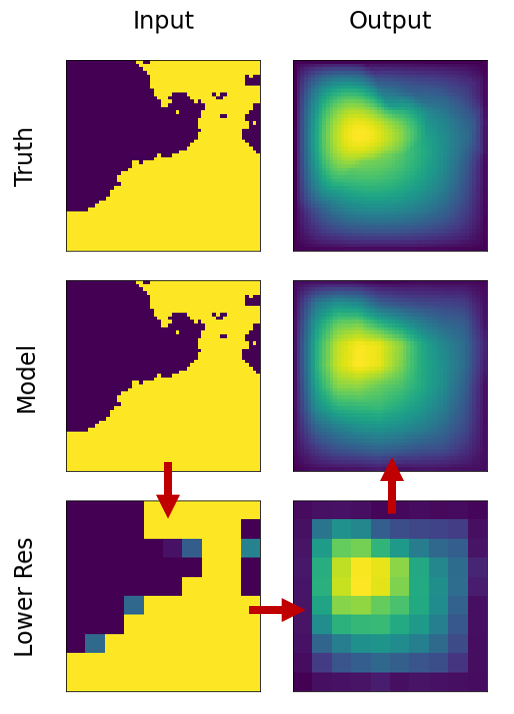}
    \caption{\textbf{Spatial resolution reduction on Darcy PDE.} Example of a scaled FNO model on Darcy data. 
    For clarity, we represent the full forward pass of what we describe as a ``scaled model" by the red arrows. 
    Note that for this PDE, the true output, i.e. the target, is actually quite blurry to begin with, making the substantial lowering of resolution that is applied appear less problematic. }
    \label{fig:darcyimg}
\end{figure}

\begin{figure}[h!]
    \centering
    \includegraphics[width=0.47\textwidth]{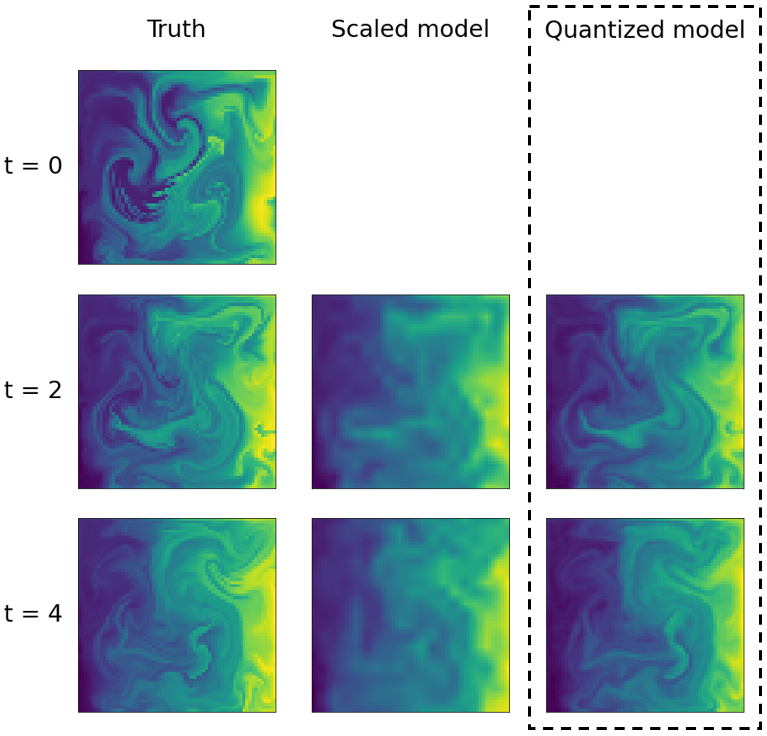}
    \caption{\textbf{Quantization retains details across time. } Example of two different UNet prediction unrollings on Navier-Stokes data. The scaled version uses a scaling factor of $0.3$, the quantized version is based on \texttt{w4a8} quantization. The left column ($t=0$) represents the last input time step used by the models, the other columns are predictions. Both models have similar inference costs, yet we see the quantized model showing more detailed predictions.}
    \label{fig:navierstokesimg}
\end{figure}

The results of our first set of experiments are presented in Figure \ref{fig:results1}, where we compare quantization against spatial scaling across models and architectures. The plots can be interpreted as follows: If the triangles (quantized models) lie below the circles (scaled models), quantization is a more robust way to reduce inference costs. We observe this is the case for all model and dataset combinations, except for the U-Net on the Darcy dataset. The reason why scaling is relatively more successful on the Darcy dataset can be understood by looking at some examples. Comparing Figure~\ref{fig:navierstokesimg} to Figure~\ref{fig:darcyimg}, the Darcy targets are already quite blurry. Generally, the dataset appears less sensitive to details in the input, making it potentially less likely that important information is lost in the process of changing spatial resolution in a forward pass.


\begin{figure*}[ht]
    \centering    
    \begin{subfigure}[b]{\textwidth}
        \includegraphics[width=\textwidth]{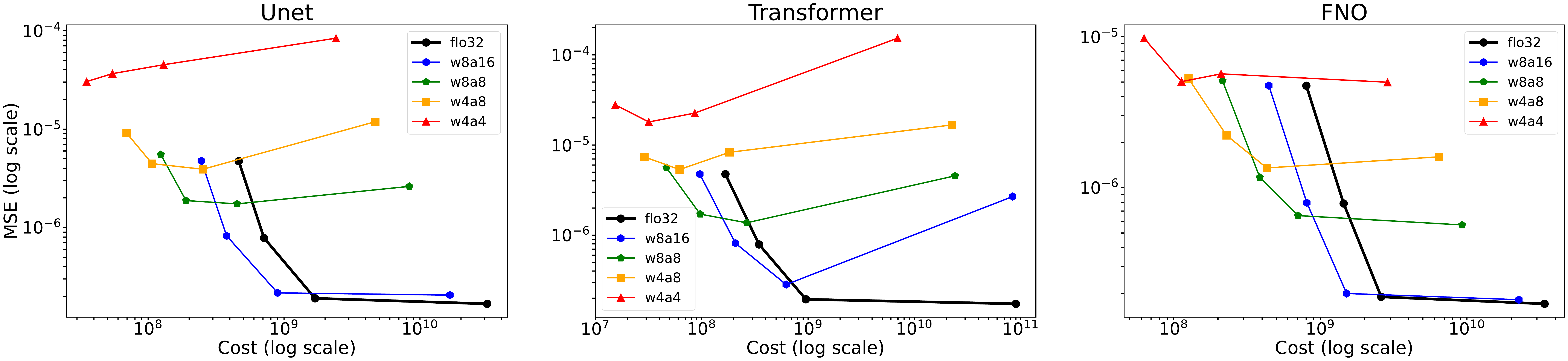}
        \caption{\textbf{1D Diffusion-Sorption Equations.}\label{fig:diffsorpplot3}}
    \end{subfigure}
    \vspace{1em}
    \begin{subfigure}[b]{\textwidth}
        \includegraphics[width=\textwidth]{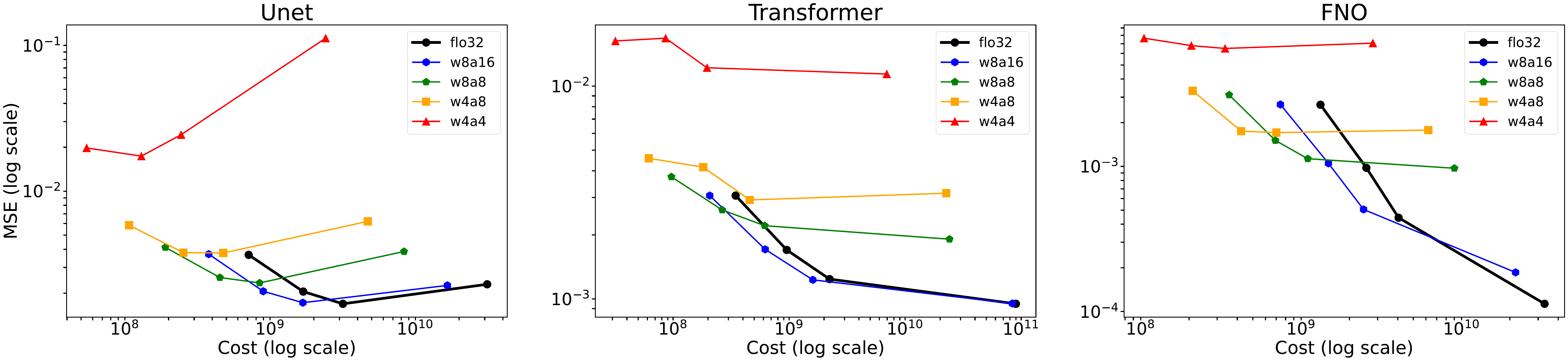}
        \caption{\textbf{1D Burgers' Equations.}\label{fig:burgersplot3}}
    \end{subfigure}
    \vspace{1em}
    \begin{subfigure}[b]{\textwidth}
        \includegraphics[width=\textwidth]{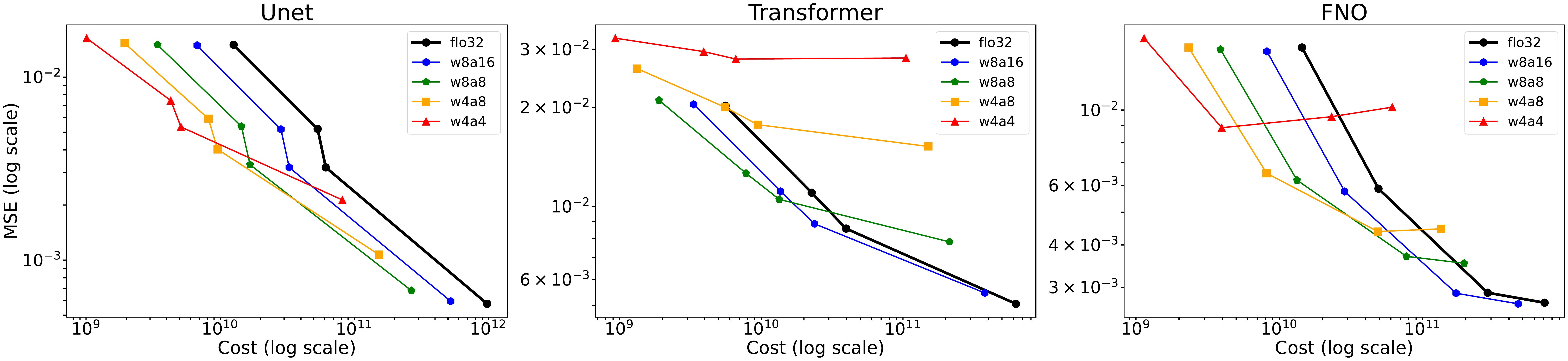}
        \caption{\textbf{2D Navier-Stokes Equations.}\label{fig:navierstokesplot3}}
    \end{subfigure}
    \vspace{1em}
    \begin{subfigure}[b]{\textwidth}
        \includegraphics[width=\textwidth]{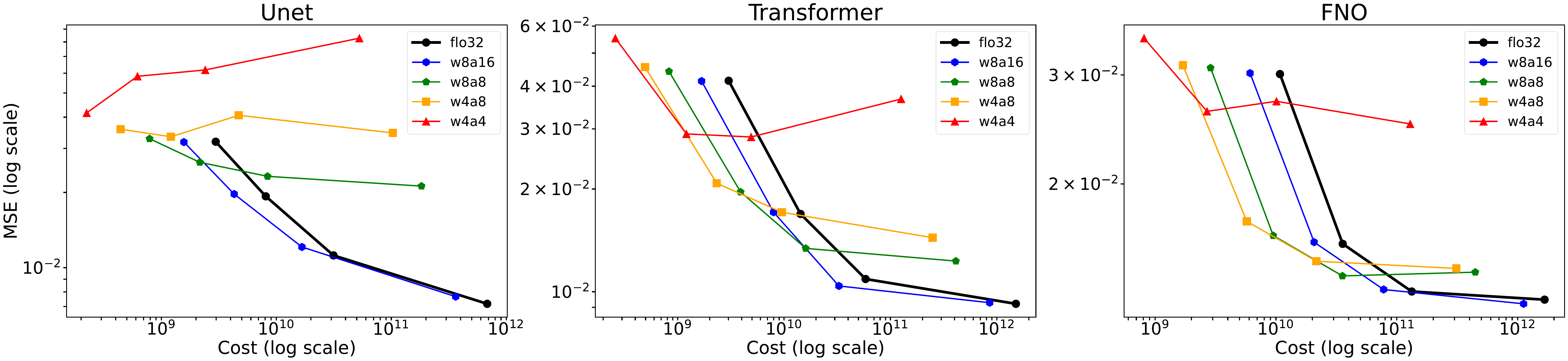}
        \caption{\textbf{2D Darcy Equations.}\label{fig:darcyplot3}}
    \end{subfigure}

    \caption{\textbf{Optimal weight and activation quantization levels yield Pareto-optimal performances.} We vary quantization levels and spatial scaling factors and observe Pareto-optimal performances compared to non-quantized (gray) models. The four points on each line correspond to different scaling levels; the right-most point is the original resolution.
    }
    \label{fig:results2}
\end{figure*}

\begin{figure}[t]
    \centering
    \includegraphics[width=0.47\textwidth]{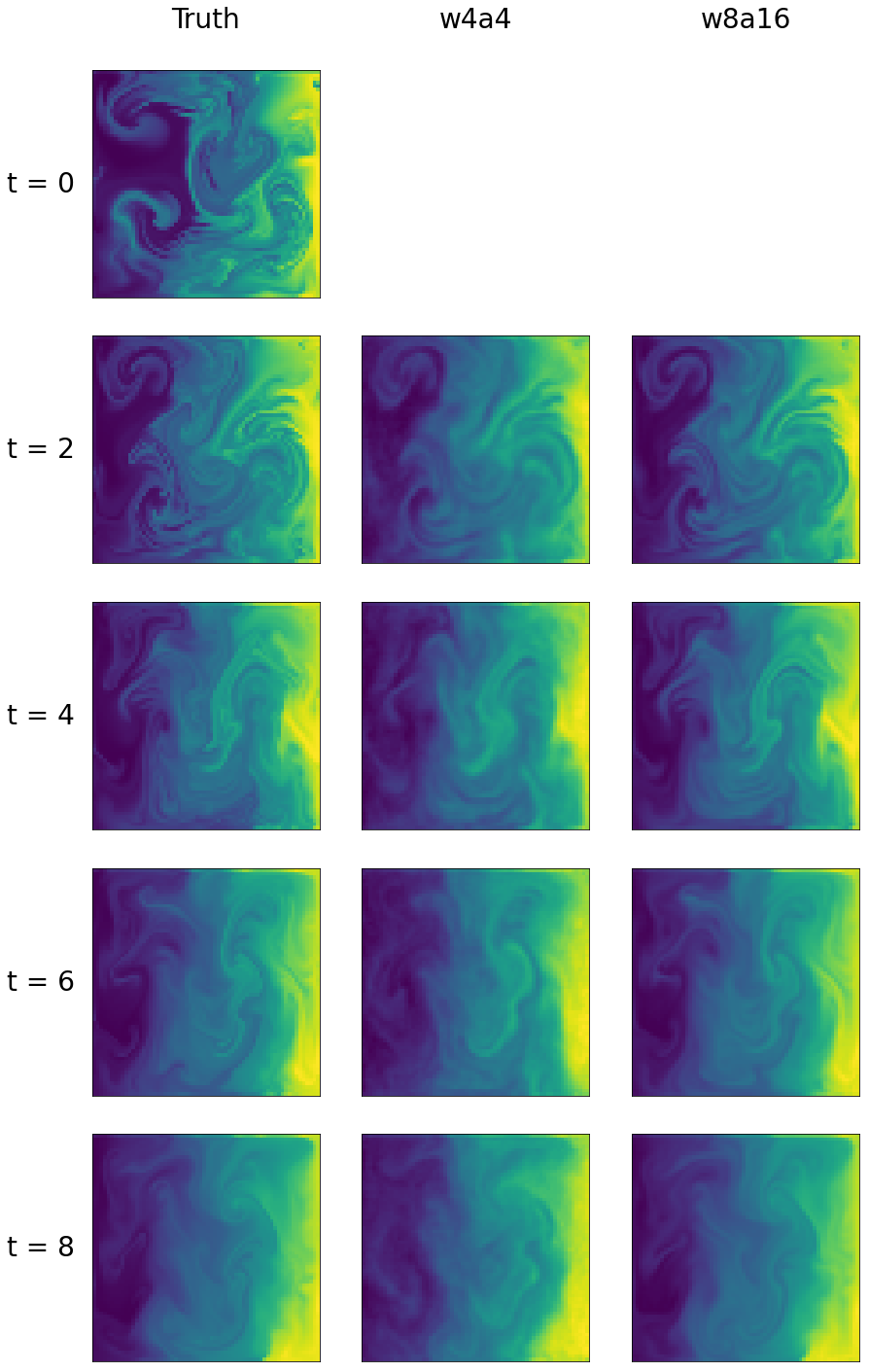}
    \caption{\textbf{Two differently quantized models unrolled for more timesteps.} Example of two different UNet prediction unrollings on Navier-Stokes data.}
    \label{fig:navierstokesimg2}
\end{figure}

\subsubsection{Combining quantization and scaling}

The datasets we use have been generated in a higher resolution than is actually used in the experiments. Note that this is common practice: the numerical solvers are applied to a much finer grid to make sure that the data is a relatively accurate representation of the underlying PDE \cite{PDEArena1, PDEBench, FNO}.
However, we now consider the hypothetical situation where one is indeed interested in minimizing loss on the higher resolution. 
Results for these experiments are found in Figures \ref{fig:results2}.

First, we find that when the original resolution is relatively high, reducing it may not affect performance very much for some of these datasets (\textit{e.g.}, Figure~\ref{fig:results2}a), and yields a good reducing in compute cost. However, as more scaling is applied, we note that switching to a lower-bitwidth quantization regime quickly becomes better than scaling alone.
In particular, we find that across almost all PDEs and neural PDE solvers, the cost vs error Pareto-optimal curve is achieved by quantization.
In other words, we find that quantization is another ``knob'' that one can turn to reduce costs, besides only changing resolution. Not only is this ``knob'' available, but the results show that is essential for optimal efficiency.
Note that the particular choice of how much quantization and how much scaling ought to be applied depends on the operating point, that can, for example, be specified by a lower bound on the MSE.

\section{Discussion and conclusion}
We have shown that quantization can be an effective solution to make neural PDE-solvers more cost-efficient. In particular, it can achieve better results in terms of prediction loss vs inference cost trade-off than the conventional method of making computational cost manageable, which is reducing spatial resolution. 
To some extent the results depend on the type of data at hand: If the original data is of extremely, and perhaps unnecessarily, high resolution, there may be at first a clear benefit to be obtained by simply reducing spatial resolution. However, we similarly find that there is no increase in loss by quantizing network weights and activations to \texttt{Int16}, with weights even quantized to \texttt{Int8} in some cases.
We note that both cost-reducing methods, quantization and reducing spatial resolution, have a certain breaking point, depending on the model and dataset at hand, at which their effect on prediction loss increases rapidly. Our results indicate that when one of these methods reaches its breaking point, one can further improve on the loss vs inference trade-off by continuing with the other method. To achieve Pareto-optimal solutions it is necessary to apply both methods simultaneously.

In this paper, we have manually selected levels of quantization and resolution scaling, showing empirically the importance of quantization as a new knob that can be turned to achieve better results for similar inference cost in existing models. However, in future research, we will investigate how a neural PDE-solver network can automatically detect the appropriate scaling or quantization levels to optimize the prediction loss vs inference cost trade-off.
Another direction for future research is to experiment with real-world weather and climate prediction data. Although our current research is based on toy data, we actually expect the results to be even more pronounced when working with real-world data: The recurring theme (see for instance \cite{randall2007climate,scale1,scale2}) is that insufficient computational power inhibits the use of higher data resolutions which would solve many modeling problems and prediction inaccuracies. We propose quantization as a strategy to free up computational resources, allowing models to be defined on these desired higher resolutions.

\subsection*{Acknowledgements}
We thank SURF for the support in using the National Supercomputer Snellius.
This work is financially supported by Qualcomm Technologies Inc., the University of Amsterdam and the allowance Top consortia for Knowledge and Innovation (TKIs) from the Netherlands Ministry of Economic Affairs and Climate Policy.

\clearpage
{\small
\bibliographystyle{ieee_fullname}
\bibliography{bibliography.bib}
}

\clearpage
\appendix
\section{Appendix}

\subsection{PDEs and Data} \label{appendix:datasets}
We train our neural networks on datasets that contain solutions $\bm{u}$ for different boundary and initial conditions so that it is able to generalize across these conditions, without the need for retraining. Datasets are obtained as follows:
\begin{itemize}
\item Generate a pair of initial conditions $\bm{u}^0(\bm{x})$ and boundary conditions $\mathcal{B}[\bm{u}](t,\bm{x})=0$ and evaluate these values on the relevant subsets of our grid $\mathcal{T}\times \mathcal{X}$.
\item Use a conventional high-accuracy numerical solver to obtain $\bm{u}(t,\bm{x})$ for all $(\bm{x},t) \in \mathcal{T}\times\mathcal{X}$.
\item Pick a series of input indices, and subsequent target indices, from the time interval $\mathcal{T}$, starting from a possibly randomly chosen location. The values of $\bm{u}(\cdot,\bm{x})$ at these indices will form the inputs and targets in our training scheme. 
\end{itemize}

\paragraph{Burger's equation}
The Burger's equation is a common PDE that arises in fluid dynamics and nonlinear wave phenomena. In 1D the PDE, given the domain that we use, is given by
\begin{eqnarray}
\frac{{\partial u}}{{\partial t}} + u\frac{{\partial u}}{{\partial x}} = \frac{\nu}{\pi} \frac{{\partial^2 u}}{{\partial x^2}}, \; \; x\in(0,1) \; t\in(0,2]
\end{eqnarray}
where $u$ represents the speed of the fluid at a certain place and time, and $\nu$ is the viscosity coefficient. The Burger's equation describes the conservation of mass and momentum in a one-dimensional fluid flow, taking into account both convection effects (\(u\frac{{\partial u}}{{\partial x}}\)) and diffusion effects (\(\nu \frac{{\partial^2 u}}{{\partial x^2}}\)). 

We use the 1D Burger's equation dataset from \cite{PDEBench}. It is defined with a spatial resolution of 1024, with periodic boundary conditions, and temporal resolution of 200. The dataset consists of 9000 train and 1000 test trajectories started from samples of different initial conditions that are formed using a superposition of randomly chosen sinusoidal waves. A viscosity coefficient of $\nu=0.001$ is used.\\

\paragraph{Darcy's Law}
The steady state 2D Darcy flow equation is a partial differential equation (PDE) that describes the flow of fluid through a porous medium. We use the PDE and domain expressed as

\begin{eqnarray}
-\nabla \left( a(x) \nabla u(x) \right) &=& f(x), \; \; x\in(0,1)^2, \\ \nonumber
u(x)&=&0, \; x \in \partial(0,1)^2,
\end{eqnarray}

where \(a(x)\) is a diffusion coefficient based on the permeability of the porous medium and the dynamic viscosity of the fluid, \(u(x)\) represents the pressure of the fluid, and \(f\) represents any external sources or sinks of fluid within the domain. We set \(f\) to constant 1 and train an operator that maps $a(x)$ to the solution $u(x)$.\\
We use the Darcy flow dataset from \cite{FNO}. It is defined on a spatial grid of $421\times 421$. We use 1024 train elements ($(a(x),u(x))$ pairs) and 100 validation elements.   Details for how $a(x)$ is randomly generated for each data element can be found in \cite{GalerkinTransformer}.

\paragraph{Navier-Stokes equation}
The 2D Navier-Stokes Equation and domain that we use for our experiments is given by
\begin{eqnarray}
    \frac{{\partial \bm{v}(\bm{x},t)}}{{\partial t}} &=& -\bm{v}(\bm{x},t) \cdot \nabla \bm{v}(\bm{x},t) + \nu \nabla^2 \bm{v}(\bm{x},t)\\ \nonumber &-& \nabla p(\bm{x},t) +\bm{f}(\bm{x}), \; \; x\in(0,32)^2, \; t\in(0,21].
\end{eqnarray}
It describes the flow of a fluid in terms of its velocity components $\bm{v}$, the viscosity $\nu$, and a buoyancy term $\bm{f}$. We assume incompressibility, so \(\nabla \cdot \bm{v} = 0 \), and Dirichlet boundary conditions ($\bm{v}=0$).  \\
The dataset is taken from \cite{PDEArena1}. A viscosity of $\nu = 0.01$ is used, and a buoyancy factor of $\bm{f}=(0,0.5)^T$. While generating the data, the pressure field $p$ is solved first, before subtracting its spatial gradients. In addition to the two velocity field components a scalar field $s(\bm{x})$ is introduced that is being transported through the velocity field. Its evolution is determined by
\begin{eqnarray}
    \frac{\partial s}{\partial t} = -\bm{v}(\bm{x},t) \nabla s,
\end{eqnarray}
with Neumann boundaries $\frac{\partial s}{\partial \bm{x}} = 0$ on the edge of the domain.
For more details, see \cite{PDEArena1,PDEArena2}. The full dataset consists of 2080 train samples and 1088 test samples.

\paragraph{Diffusion-Sorption Equation}
The diffusion-sorption equation models a diffusion process that is retarded by a sorption process. The 1D PDE is given by:
\begin{eqnarray}
\frac{{\partial u}}{{\partial t}} = D/R(u) \frac{{\partial^2 u}}{{\partial x^2}} \; \; x\in(0,1) \; t\in(0,500],
\end{eqnarray}
where $D=0.0005$ is the effective diffusion coefficient, and $R(u)=1+2.16u^{-0.126}$ is the retardation factor hindering the diffusion process.  This equation is applicable to, for example, groundwater contaminant transport.\\
The boundary conditions are $u(t,0)=1$ and $u(t,1)=D\frac{\partial u}{\partial x} (t,1)$
The dataset, taken from \cite{PDEBench}, is discretized into 1024 spatial steps and 501 time steps. There are 9000 train trajectories and 1000 test trajectories, each based on different randomly generated initial conditions using $u(0,x) \sim U(0, 0.2)$ for $x \in (0, 1)$.

\subsection{Hyperparameter specifications}\label{appendix:params}
We summarize the hyperparameters used per dataset in Table \ref{table:hparams}.

\begin{table}[h!]
\begin{tabular}{lrrrr}
\toprule
Hyperparams & DiffSorp           & Burgers'              & N.S.             & Darcy  \\
\midrule
Epochs            & $200$       & $200$       & $100$           & $400$     \\ 
QAT Epochs       & $100$       & $50$         & $50$            & $100$    \\ 
Batch size        & $50$        & $50$         & $16$            & $4$      \\
Learning rate     & $1$e-$3$      & $1$e-$3$       & $1$e-$3$          & $5$e-$4$   \\
Weight decay      & $1$e-$6$      & $1$e-$6$       & $1$e-$6$          & $1$e-$6$  \\ 
QAT learn. rate & $1$e-$4$      & $1$e-$4$       & $1$e-$4$          & $1$e-$4$   \\ 
Input steps   & $5$         & $5$          & $4$             & $1$       \\ 
Output steps  & $5$         & $5$          & $1$             & $1$      \\ 
Train steps   & $10$        & $20$         & $1$             & $1$      \\
Test steps    & $10$        & $20$         & $1$             & $1$      \\
Subsample $t$     & $2$         & $5$          & $1$             & $1$       \\ 
Subsample $x$     & $32$     & $16$     & $2$          & $8$   \\
\bottomrule
\end{tabular}
\caption{\label{table:hparams} Dataset-related hyperparameters for all models per experiment. The steps refer to consecutive time steps for the time-dependent PDEs, while the Darcy PDE can optionally be interpreted as having inputs at $t=0$ and outputs at $t=1$.}
\end{table}
In the second session of experiments we did not subsample the spatial grid, except for the Darcy dataset for which we subsampled every $2$ grid points. The first three scaling levels applied in figure \ref{fig:results2} correspond (from left to right) to $0.01, 0.02, 0.05$ for the DiffSorp data, $0.02, 0.05, 0.1$ for the Burgers data, $0.1, 0.2, 0.5$ for the Navier-Stokes data and $0.05, 0.1, 0.2$ for the Darcy data.
The loss measure used in all datasets is the MSE as described in equation \ref{equation:loss}. However, for the Darcy dataset, we also normalize each element in the sum by dividing by the squared targets, and take the squared root of the resulting sum.

The UNet is taken from \cite{PDEArena1}, but in order to make its size comparable to the other models we use 16 hidden channels for the Navier-Stokes dataset and 8 hidden channels for the other datasets. The FNO model is taken from \cite{FNO}, using 4 layers, a width of 128, 32 modes for the 2D datasets, and 16 modes for the 1D datasets. The Transformer is taken from \cite{GalerkinTransformer}. It uses 6 encoder layers, 128 hidden channels and a Galerkin attention type for the 2D datasets, and 4 encoder layers, 32 hidden channels and Fourier attention type for the 1D datasets.

\subsection{Inference Cost Calculation Details} \label{appendix:inferencecost}
We describe a few differences compared to the regular deepspeed library \cite{deepspeed}. Most standard deep learning operations rely on big matrix multiplications and as such deepspeed outputs the number of MACs used in their corresponding modules. On the other hand, there are some operations that have no MACs, and deepspeed simply outputs the number of FLOPs. However, because we care to differentiate addition and multiplication operations for our proxy measure of inference cost, we add some manual changes to deepspeed so that multiplications are properly accounted for.
\begin{itemize}
    \item We assume bilinear interpolation to be three times the cost of linear interpolation, and for linear interpolation we assume $2$ multiplications and $4$ additions per output point.
    \item The FNO model uses Fast Fourier Transforms, which are not encountered for in deepspeed. To be able to take these into account in our proxy for model inference cost we assume a complexity of $N  \lceil \log_2N \rceil$ (additions and multiplications), which we divide by $2$ when the real-valued FFT is used.
    \item In deepspeed no MACs are assigned to the einsum operator. Although it can in theory represent various different types of computations, in our code we only use it for basic matrix multiplications (in the FNO model). We thus change the deepspeed output accordingly.
\end{itemize}
\end{document}